\documentclass[11pt]{article}

\usepackage{acl}
\usepackage{graphicx}
\usepackage{latexsym}
\usepackage[T1]{fontenc}
\usepackage{tgheros}
\usepackage{tgcursor}

\usepackage[T1]{fontenc}

\usepackage[utf8]{inputenc}

\usepackage{microtype}

\title{HealthcareNLP: where are we and what's next?}

\author{ Lifeng Han, Paul Rayson,  Andrew Moore, Goran Nenadic, \and Suzan Verberne   \\ On behalf of the 4D PICTURE consortium and beyond \\
\texttt{Leiden University Medical Centre, NL} \\
\texttt{Lancaster University, UK} 
\texttt{University of Manchester, UK}  \\
 \texttt{The Leiden Institute of Advanced Computer Science, NL} \\
\{l.han, s.verberne\} @liacs.leidenuniv.nl \\ Tutorial presented at LREC 2026 
 \\
 \\
}
\begin{document}
\maketitle
\section*{Abstract}
This tutorial focused on Healthcare Domain Applications of NLP, what we have achieved around HealthcareNLP and the challenges that lie ahead for the future.
Existing reviews in this domain either overlook some important tasks, such as synthetic data generation for addressing privacy concerns, or explainable clinical NLP for improved integration and implementation, or fail to mention important methodologies, including retrieval augmented generation and the neural symbolic integration of large language models (LLMs) and knowledge graphs (KGs).
In light of this, the goal of this tutorial is to provide an overview introduction of the most important sub-areas of a patient and resource/data-oriented
HealthcareNLP, with \textbf{three layers} of hierarchy: 
1) \textbf{data}/resource layer: annotation guidelines, ethical approvals, governance, synthetic data; 
2) \textbf{NLP}-Eval layer: NLP tasks NER, relation extraction (RE), sentiment analysis, and linking/coding with categorised methods, leading to explainable HealthAI;
3) \textbf{patients} layer: Patient Public Involvement and Engagement (PPIE), health literacy, translation, simplification, and summarisation (\textit{aka} NLP tasks), shared decision making support.
A hands-on session or group discussion will be included in the tutorial for audiences to use HealthcareNLP applications or engage in discussing the challenges and issues in the field.
The target audiences include NLP practitioners in the healthcare application domain, NLP researchers who are interested in domain applications, Healthcare researchers (HealthcareNLP users), and students from NLP fields.
The type of tutorial is ``Introductory to CL/NLP topics (HealthcareNLP)'' and audiences do not need prior knowledge to attend this. Tutorial materials are available at \url{https://github.com/4dpicture/HealthNLP} 

\section{Introduction}
\begin{figure*}[t]
  \includegraphics[width=1\textwidth]{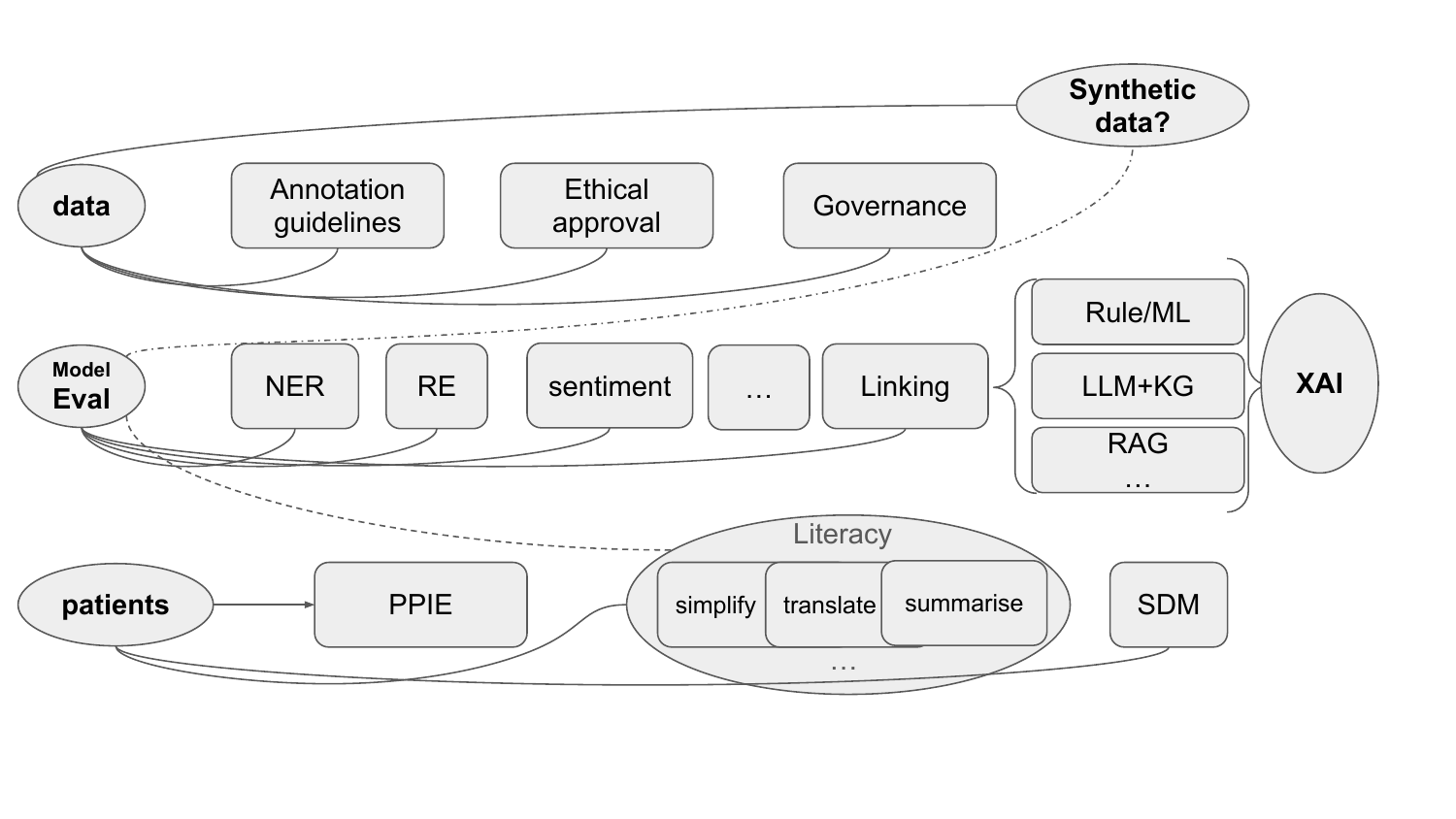}
  \caption{An overview of HealthcareNLP tasks (up-down): 1) \texttt{data-layer} annotation guidelines, ethical approvals, governance, and synthetic data applications; 2) \texttt{NLP-Eval-layer} tasks on named entity recognition, relation extractions, sentiment analysis, entity linking, and possible NLP techniques including traditional rule/ML, LLMs with KG, retrieval-augmented generation, explainable AI (XAI); 3) \texttt{patients-layer} related event: public patient involvement and engagement; text simplification / lay-language adaptation and summarisation for public health literacy, and machine translation for multilingual setting (L2 learners in a new country), shared decision making (SDM) communication support. }
  \label{fig:healthcareNLP-diag}
\end{figure*}

With the increasing capabilities of current state-of-the-art NLP models, NLP applications have been one of the hot topics in AI for social good, which includes their application in the healthcare domain.
It is not a brand new topic, since there have been NLP researchers trying to deploy NLP techniques into the biomedical domain a long time ago, and the BioNLP workshop has been held since 2002 for 23 years \footnote{\url{https://aclweb.org/aclwiki/BioNLP_Workshop}}.
There is indeed growing interest from both researchers and practitioners, such as the recently emerged workshops including CL4Health at LREC-COLING2024 and NAACL2025 \cite{cl4health-ws-2025-1}, 6th HealthNLP at ICHI2023 \footnote{\url{https://www.healthnlp.info/}}, Big Data Analytics in Healthcare at IEEE-BigData2023 \footnote{\url{https://onlineacademiccommunity.uvic.ca/healthcare4bigdata4analytic/}}, ClinicalNLP\footnote{\url{https://aclanthology.org/venues/clinicalnlp/}} etc.
new shared tasks such as BioLaySumm  \cite{xiao2025overview}, PLABA (plain language adaptation of biomedical abstract) \cite{ondov2025lessons}, etc.

This tutorial starts with an explanation of each sub-task into three layers (data resource, NLP, and patients), how they are defined, and what NLP approaches were used to address such tasks in history, up to the state-of-the-art models.

First, in the \textbf{resource} layer, we talk about 1.1) Annotation Guidelines: how the clinical documents shall be annotated and anonymised to protect privacy and address ethical concerns. 1.2) Ethical Approval and Data Governance: how data should be hosted and how access is granted for research purposes. 1.3) Synthetic data generation (connecting to NLP): how this task can be useful to address data sharing and scarcity issues. 
Second, in the \textbf{NLP}-eval layer, these tasks include 2.1) standard named entity recognition (NER) in biomedical and clinical domains such as disease, symptoms, diagnoses, treatments, and progression entity categories. 2.2) relation extraction (RE) such as drugs and side-effects, symptoms and treatments, treatments and temporality, etc. 2.3) sentiment analysis of NLP models on patient-centred text data, including patient interview transcripts, online forums, and consultations.
2.4) Clinical Coding (normalisation/linking), which automates the traditional human coders work by linking the clinical entities into clinical knowledge graphs and dictionaries; 2.5) Machine Translation (MT): on translating clinical text from one language to others to address low-resource eHealth model training obstacles as well as in a multilingual consultation environment where the patients cannot speak/understand the dominant language; 2.6) Text Simplification and Summarisation: to help patients better understand the clinical evaluation documents and healthcare information websites. 
We will introduce the method applied to these tasks from historical rule-based, to statistical, and nowadays neural network models with attention mechanisms. If time allows, we will cover the state-of-the-art retrieval-augmented generation (RAG), large language models with knowledge graphs (LLM+KG), and prompt engineering, as shown in Figure \ref{fig:healthcareNLP-diag} in the overview of HealthcareNLP picture.
Third, in the \textbf{patient} layer, we also mention some engaging events in this field, including 3.1) public and patient involvement and engagement (PPIE) practice examples from the UKRI and EU projects, 3.2) patient health literacy, and 3.3) how to better support the communication between clinicians and patients for better shared decision making (SDM).

In the end, we point out some promissing research directions following these three layers and overall HealthcareNLP fields (Fig. \ref{fig:NLP-2application-mapping-diag}).

\section{Target Audience}
This tutorial will be developed for NLP researchers who are interested in applying their models and methods to the healthcare domain, which potentially can have a real impact on social lives and healthcare systems, making them more efficient and accurate.
It is also useful for healthcare practitioners and data scientist in the hospital who go to the NLP conference to learn more knowledge on how the models they use are developed for healthcare.
For UG and PG students who are interested in AI and NLP applications, this tutorial can also be of interest as an introductory level to AI for the healthcare field.
Prior knowledge is not required to attend this event. By the end of the tutorial, the audiences shall acquire an overview of HealthcareNLP and CL4Health, what tasks are there, what models are being used, and what are the current challenges.

\section{Outline}
The tutorial aims to be 4 hours long including 30 minute break in between.
The covered topics include the following:

\begin{itemize}
    \item Data resources: annotation guidelines, anonymisation, ethical approval, governance, synthetic data usage
    \item HealthcareNLP Tasks and Evaluations: NER, RE, Linking, Sentiment Classification, Simplification, Translation, Summarisation, QA 
    \item Models applied: rule, statistical, neural models, LLM+KG, prompting, RAG, Explainable and Interpretable AI
    \item Patients: Public patient involvement and engagement (PPIE), Patient health literacy, Shared Decision Making (SDM) \cite{stiggelbout2015shared,frosch1999shared}
    \item Real-world applications of HealthcareNLP projects and models in the UK / NL, and EU project 4D PICTURE consortium \url{https://4dpicture.eu/}
    \item Hands-on experience of using some healthcareNLP platforms and applications, including our own. 
\end{itemize}

\section{Technical Requirements} 
Our technical expectations are that participants will have access to a good internet connection for online code and data experiments. 

\section{Diversity Considerations} 
The tutorial includes an MT sub-topic on translating healthcare domain data into low-resource languages so that researchers can help to build Healthcare AI/NLP systems for low-resource communities.

\section{Reading List}
Healthcare and Clinical NLP   \cite{jerfy2024growing,nazi2024large,elvas2025natural}.

\section{Tutorial Organisers/Presenters}

\textbf{Dr. Lifeng Han} is currently a researcher on the 4D Picture project \url{https://4dpicture.eu/} researching the cancer patient journey support. 
He had his first postdoctoral position with the University of Manchester from 2022 to 2024 on HealthcareNLP funded by the UKRI.
Prior to these, he finished his PhD thesis "An investigation into multi-word expressions in machine translation" \cite{han2022investigation} from ADAPT centre, Ireland.
He initiated the joint track between two workshops MWE and ClinicalNLP at MWE2023 with EACL \cite{mwe-2023-multiword} when he was co-chairing MWE2023.
He gave a main conference tutorial to LREC2022 in person / on-site on ``Meta-Evaluation of Translation Evaluation Methods: a systematic up-to-date overview'' at Marseille, France, 2022. 
His recent publications mainly focused on HealthcareNLP in the past three years. He has 50+ peer-reviewed publications, including journal articles and NLP main conference papers.


\textbf{Professor Paul Rayson} is based in the School of Computing and Communications at Lancaster University, UK and Director of the UCREL interdisciplinary research centre which carries out research in corpus linguistics and natural language processing (NLP). A long term focus of his work is semantic multilingual NLP in extreme circumstances where language is noisy e.g. in historical, learner, speech, email, txt and other CMC varieties. Along with domain experts,  he has applied his research in the areas of metaphor in cancer narratives, dementia detection, mental health, online child protection, cyber security, learner dictionaries, and text mining of biomedical literature, historical corpora, and financial narratives. He was a co-investigator of the five-year ESRC Centre for Corpus Approaches to Social Science (CASS) which was designed to bring the corpus approach to bear on a range of social sciences. 

\textbf{Professor Suzan Verberne} is based at the Leiden Institute of Advanced Computer Science (LIACS), Leiden University, The Netherlands.
She is professor of Natural Language Processing . She works on the edge between Natural Language Processing (NLP) and Information Retrieval (IR). 
She has supervised projects involving a large number of application domains and collaborations, many of which in health and medical contexts. Suzan is internationally active in both the NLP and IR communities, and a highly-valued teacher of two courses in the master programme Computer Science.

\textbf{Dr. Andrew Moore} is a research software engineer at Lancaster University, also currently working on semantic tagging and emotion analysis within the 4D picture project, with a PhD on empirical evaluation methodology for target dependent sentiment analysis \citep{andrewmoore2021} as well as conference papers on this topic. He has worked on semantic tagging while working at Lancaster University creating, maintaining, and extending the Python Multilingual Ucrel Semantic Analysis System (PyMUSAS)\footnote{\url{https://ucrel.github.io/pymusas/}} software with Professor Paul Rayson. He has also previously worked within industry at the World Intellectual Property Organization (WIPO) working on topics such as Machine Translation Quality Estimation and semantic search. 

\textbf{Professor Goran Nenadic} is based in Department of Computer Science, University of Manchester, with expertise in HealthNLP, as co-/PI of multiple UK funded projects.
Current research projects focus on large-scale extraction and curation of biomedical information and clinical/epidemiological findings, by combing rule-based and data-driven approaches. He is also interested in processing healthcare social media.
He leads the UK healthcare text analytics network (Healtex) and was the founding Editor-in-Chief of Journal of Biomedical Semantics.

\section{Other Information}
The number of audiences who attended our tutorial this year is around 50-100. 

\section{Ethics Statement}
All the examples and corpora used in the tutorial are either public data from shared task challenges approved by the shared task organisers or anonymised data samples from our research projects.

\section{Further Reading and Resources}
\label{sec:further-reading}
\begin{figure*}[t]
  \includegraphics[width=1\textwidth]{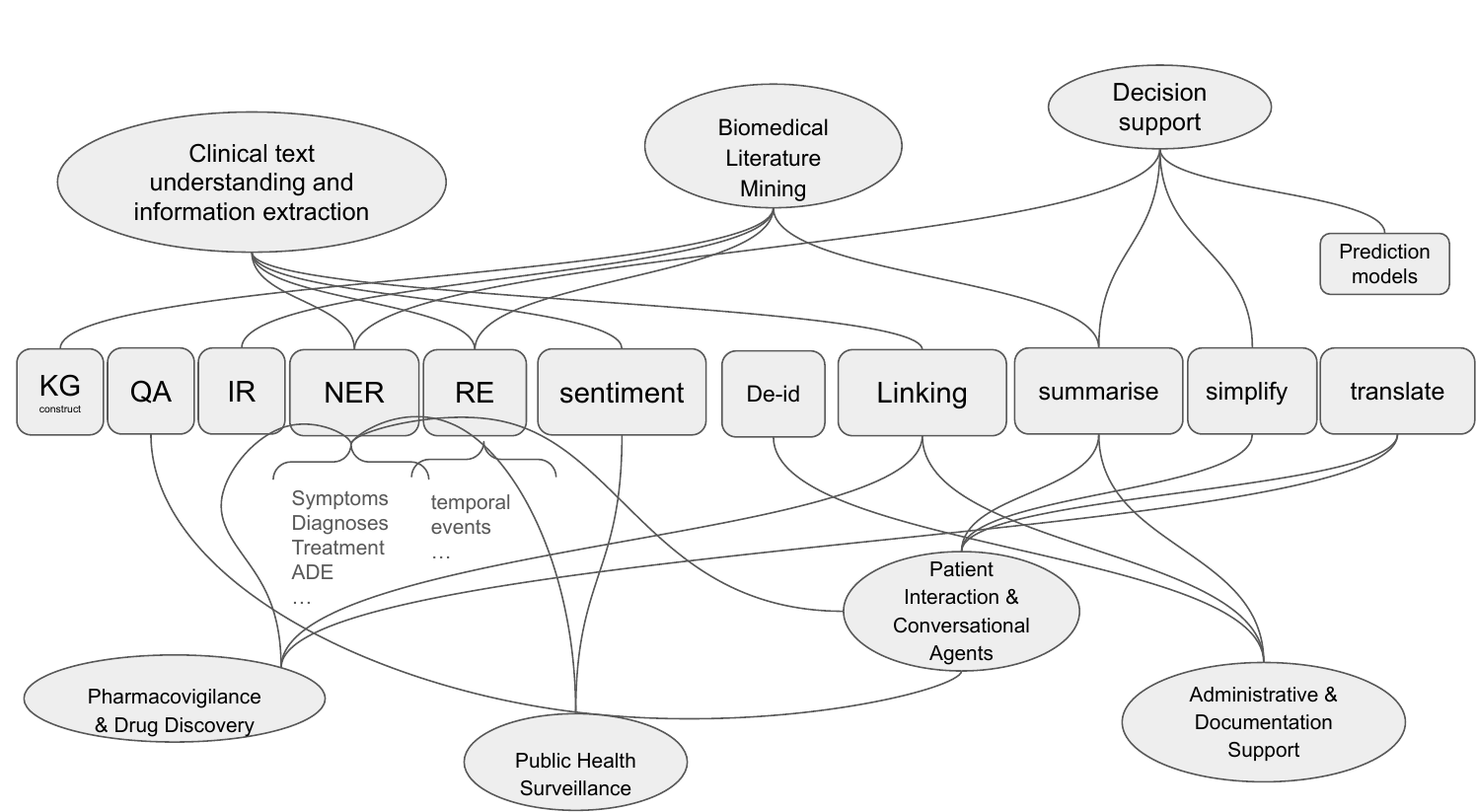}
  \caption{NLP-related tasks and their mappings to applications (Future Perspectives)
  }
  \label{fig:NLP-2application-mapping-diag}
\end{figure*}

The tutorial is accompanied by an annotated bibliography covering representative work across the data, NLP, and patient layers of Healthcare NLP. The curated bibliography is provided in Appendix~\ref{sec:annotated-bibliography}; a living version is maintained in the tutorial repository at \url{https://github.com/4dpicture/HealthNLP} including the full presentation slides of the tutorial. 
The future-perspectives mapping in Figure~\ref{fig:NLP-2application-mapping-diag} summarises connections between NLP tasks and healthcare applications.

\section{Acknowledgement}
We thank Ida Korfage, Sheila Payne, and Judith Spek for their feedback on the abstract of this work.
The 4D PICTURE: Funded by the European Union under Horizon Europe Work Programme 101057332. Views and opinions expressed are however those of the author(s) only and do not necessarily reflect those of the European Union or the European Health and Digital Executive Agency (HaDEA). Neither the European Union nor the granting authority can be held responsible for them.
The UK team are funded under the Innovate UK Horizon Europe Guarantee Programme, UKRI Reference Number: 10041120.\\
GN is supported by the grant “Assembling the Data Jigsaw: Powering Robust Research on the Causes, Determinants, and Outcomes of MSK Disease”, and the grant “Integrating hospital outpatient letters into the healthcare data space” (EP/V047949/1; funder: UKRI/EPSRC).

\appendix
\section{Annotated Bibliography for Healthcare NLP}
\label{sec:annotated-bibliography}

This appendix provides an annotated bibliography accompanying the tutorial ``\textit{Healthcare NLP: Where Are We and What Is Next?}''. It is organised around the data, NLP, and patient layers used throughout the tutorial. A living version is maintained in the tutorial repository at \url{https://github.com/4dpicture/HealthNLP}. The entries below provide short annotations to papers and resources discussed in or closely related to the tutorial. The appendix is intended as a \textit{companion reading guide} rather than a systematic review. 


\subsection{Introduction: core papers}

\citet{wu2022survey} survey clinical NLP projects in the United Kingdom between 2007 and 2022. They show that clinical NLP publications double every five years.

\citet{locke2021natural} provide an introductory and practical review of NLP in medicine. They cover tasks from information extraction to clinical decision support and make the topic accessible to a clinical audience.

\citet{spasic2020clinical} is a data-centric systematic review of clinical text datasets used in machine learning between 2010 and 2016. They provide an overview of corpora and annotation practices across specialties.

\citet{elvas2025natural-review} carry out a scoping review of NLP in medical text processing (2019--2023). They cover tasks from classification to information extraction and report work in multiple languages beyond English.

\citet{maddox2025generativeAI} summarise the progress generative AI has made in healthcare and the challenges that remain, as assessed by a National Academy of Medicine workshop of experts.

\subsection{Data layer}

\subsubsection{Annotation}

\citet{gurulingappa2012} set a good example of a text annotation process in the healthcare domain: they developed a benchmark corpus of nearly 3,000 case reports manually annotated with drugs and adverse effects, showing that guideline development should be a cyclic process guided by inter-annotator agreement until convergence.

An instance of a sophisticated guideline is proposed by \citet{Schulz2023TowardsPO} to use the semantic standards SNOMED CT and HL7 FHIR. They design a hierarchical annotation schema with \textit{core} and \textit{sub}-concepts based on ontological principles, enriched with A-Box and T-Box annotations using symbolic and neural reasoning.

NER: \citet{yada-etal-2020-towards} aimed at a versatile annotation guideline using critical lung disease data. Their design avoids heavy medical-knowledge requirements by focusing on linguistic patterns, annotating entity types including diseases, anatomical entities, features, temporal expressions, medicines, and clinical context.

\citet{xiaclinical2012} discuss the challenges and strategies of clinical corpus annotation, emphasising the importance of involving NLP researchers early alongside clinical experts.

Non-English examples:
\begin{itemize}
\item SemClinBr by \citet{oliveira2022semclinbr} for Portuguese electronic health records: a multi-institutional, multi-specialty semantically annotated corpus.
\end{itemize}
\begin{itemize} 
\item \citet{miranda2020overview} for Spanish clinical coding systems of medical documents, presented at the CodiEsp track of CLEF eHealth 2020.
\end{itemize}
\begin{itemize}
\item CARDIO:DE by \citet{richter2023distributable}, a distributable German clinical corpus of cardiovascular routine doctor's letters.
\end{itemize}
\begin{itemize}
\item \citet{becker2025extending} extend CARDIO:DE with additional annotation guidelines and an evaluation of NLP approaches for clinical applications.
\end{itemize}

\subsubsection{Ethical Concerns}

\citet{suster2017short} provide a short review of ethical challenges in clinical NLP, discussing data sanitisation, consent, secure access, and sources of less sensitive data; they also encourage reporting of model bias to avoid socially harmful impact.

\citet{benton-etal-2017-ethical} outline ethical research protocols for social media health research, emphasising the roles of Institutional Review Boards (IRBs), informed consent, de-identification, and the risks of data linking.

\citet{lewis-etal-2017-integrating} address the integration of personal data protection with open science in research ethics, offering practical guidance for the NLP community.

\citet{10.1093/jamiaopen/ooac039Walk2022} present a scoping review of ethics considerations in clinical NLP, analysing 22 papers through the lens of bias and fairness, with attention to the ML development pipeline (design, data, algorithm, critique).

\citet{khattak2023ethical} discuss ethical considerations and challenges in deploying NLP systems in healthcare. They cover data privacy, security, model bias, medical liability, explainability, and transparency.

\citet{fu2023recommended} conduct a scoping review of recommended practices and ethical considerations for NLP-assisted observational research, emphasising the WHO's six principles for ethical AI use in health: protecting autonomy, promoting human well-being, ensuring transparency, fostering accountability, ensuring inclusiveness, and promoting sustainability.

\citet{bak2025ethical} report on the embedded ethics review of our EU-funded 4D PICTURE project on cancer patient quality-of-life support, introducing ideas such as an embedded ethicist position, psychological safety in research teams, and anticipating techno-moral change.

\subsubsection{Health Data Governance}

\citet{jones2020toward} conduct a UK network study on developing data governance standards for reusing clinical free text in health research. They identify barriers around authoritative guidance, public transparency, de-identification efficacy, and the use of accredited data safe havens.

\citet{di2021assessing-EU} survey data protection and governance in health information systems across 15 centres in 10 European countries. They find high variability in privacy practices and proposing a Privacy and Ethics Impact and Performance Assessment (PEIPA) methodology.

\citet{wang2022big} describe the big data platform of West China Hospital (online since 2020). They cover more than 12 million patients, 75 million visits, and 8,475 data variables. This is an example of large-scale clinical data governance in practice.

\citet{lopez2023clinical} present a Rare Cancer Data Ecosystem for the European healthcare setting, using emerging interoperability technologies and AI to improve data quality and organisation for rare cancer patient care.

\citet{arigbabu2024data-nightingale} study data governance in AI-enabled healthcare via a survey of 843 users. They find a positive link between awareness of AI projects and trust in healthcare providers, and stress the importance of transparent communication about data use.

Promising technical approaches for safer data access include {federated learning} (\citet{xu2021federated}; \citet{nguyen2022federated}; \citet{nguyen2022federated}) and {access-level control} (\citet{abouelmehdi2018big-access}; \citet{xiang2021privacy-access}).

\subsubsection{Use of Synthetic Data}

\citet{el2024evaluation} evaluate the replicability of analyses using synthetic health data, a useful reference for the methodology of tabular synthetic data generation in the health domain.

\citet{belkadi2025lt3} introduce LT3 (Label-To-Text Transformer), a model that generates synthetic medical prescriptions conditioned on label vocabularies, enabling NER training without access to real patient data. Models trained on the synthetic data achieve 96--98\% F1 on the n2c2-2018 medication extraction task.

\citet{ren2025malei} present \textbf{Synthetic4Health}, a pipeline for generating annotated synthetic clinical letters from structured knowledge, evaluated for lexical similarity, utility (train-on-synthetic, test-on-real), and diversity.

\citet{kattenberg2025generating} generates a synthetic Dutch medical dataset of episode descriptions using a fine-tuned GPT-2 model conditioned on ICPC class labels. He demonstrates reasonable classification accuracy but reduced diversity compared to real data.

\citet{belkadi2025mlm4synmed} introduce \textbf{MLM4SynMed}, using masked language modelling to generate synthetic free-text medical records, extending synthetic data generation to a more challenging and heterogeneous text type.

\subsection{NLP layer}

\subsubsection{Classification and Prediction Models}

\citet{beeksma2019predicting} use an LSTM recurrent neural network on electronic medical records to predict life expectancy, demonstrating that sequential models can assist clinicians in timely end-of-life conversations based on prognostication.

\citet{Shmatko2025Learning} present \textbf{Delphi}, a transformer-based generative model trained on large-scale UK Biobank health records to learn long-term disease progression trajectories, enabling multi-disease risk prediction over decades and capturing comorbidity patterns.

\citet{LIANG2025978AI-generated} fine-tune Llama-2-13B with AI-generated medical diagnoses for ICD coding in gynaecologic oncology, showing a practical route to using LLMs for clinical classification with limited real labelled data.

\subsubsection{Entity Recognition and De-identification}

\citet{Belkadi2023etal_PLM4clinicalNER} compare pre-trained encoder-based language models (BERT, BioBERT, ClinicalBERT) fine-tuned for drug and adverse drug effect entity recognition. They find that domain-specific models achieve similar F1 (0.84) to general BERT despite higher precision.

\citet{romero2025medication} investigate ensemble learning over eight diverse encoder-based models (including RoBERTa-L, BioMedRoBERTa, PubMedBERT) for medication extraction on the n2c2-2018 EHR dataset. Their non-BIO word-level ensemble reaches macro-F1 0.8821, while the BIO-aware max-logit voting ensemble reaches macro-F1 0.8232.

\citet{magge2021deep} present \textbf{DeepADEMiner}, a deep learning pharmacovigilance pipeline for extracting and normalising adverse drug event mentions from Twitter, combining NER with entity normalisation to a medical ontology.

\citet{dirkson2022automated} show that automated NLP extraction from online patient forums (GIST Support International on Facebook) can complement pharmacovigilance data for rare cancers, identifying drug--side-effect relations not yet in official databases.

\citet{wang2025gptner} introduce \textbf{GPT-NER}, a few-shot NER approach using generative LLMs. It cannot beat a fine-tuned model when labelled data is plentiful, but significantly outperforms supervised models when training data is extremely scarce.

\subsubsection{Biomedical Entity Linking}

\citet{liu2021sap} propose \textbf{SapBERT} (Self-Alignment Pretraining for Biomedical Entity Representations), which uses contrastive learning on UMLS synonyms to bring synonym embeddings closer together than non-synonyms, establishing a strong baseline for biomedical entity linking.

\citet{hartendorp2024} adapt SapBERT for Dutch by fine-tuning on an automatically generated Wikipedia corpus, then apply it to a Dutch cancer patient forum (kanker.nl), reporting that NER errors are currently a larger bottleneck than the linker itself.

\citet{mazzucato2026advancements} explore both NER and entity linking with decoder-only LLMs (GPT-4o) in a cross-lingual setting (Dutch and Italian EHRs, CLEF eHealth data), endorsing the potential of generative LLMs for multilingual healthcare data harmonisation.

\subsubsection{Relation Extraction}

\textit{(Left out from the tutorial)}

\subsubsection{Sentiment Analysis}

\citet{maehlum-etal-2024-difficult} add sentiment annotations to Norwegian patient-survey free-text comments and evaluate Norwegian LLMs as alternative annotators. Their experiments show that zero-shot LLMs perform well above baseline for binary sentiment, but do not match human annotators on the full dataset.

\citet{ronningstad-etal-2025-mixed} explore within- and cross-domain training for Norwegian patient feedback sentiment. They find that fine-tuned BERT and T5 models outperform zero- and few-shot LLMs on four-way sentiment, and that out-of-domain data (NoReC) can improve in-domain results when in-domain data is limited.

\citet{storset-etal-2026-pain} extend NorPaC with aspect categories aligned to patient-centred care quality dimensions and their associated sentiment labels. They find that joint training on aspect extraction and sentiment outperforms a pipeline approach.

\citet{lal-etal-2024-analysing} conduct emotion and sentiment analysis of 1,500 Reddit posts about cancer stages and treatments, using pre-trained BERT-like and lexicon-based models. They highlight the challenges of contradictory emotions and posts requiring additional context.

\subsubsection{Summarization, Simplification, and Translation}

\textit{(Left out from the tutorial)}

\subsubsection{Explainable NLP}

\textit{(Left out from the tutorial)}

\subsection{Patient layer}

\subsubsection{Privacy}

De-identification is a key prerequisite for releasing clinical NLP data. \citet{micheletti2024explorationmaskedcausallanguage} de-identify clinical texts using biomedical BERT variants with comprehensive risk assessment, contributing to the MORMOR-KARL shared task at LEGAL 2026.

Patients' attitudes towards health data sharing are explored by \citet{subramanian2024patient}, who find that HIPAA does not fully prevent data breaches and explore industry-standard alternatives. \citet{oherrin2004hipaa} examine how HIPAA regulations affect medical-record research and access to records for clinical researchers.

\subsubsection{Public and Patient Involvement and Engagement (PPIE)}

\citet{moschogianis2025feasibility} study the feasibility and added value of electronic patient-generated data (questionnaires) for people with musculoskeletal conditions, demonstrating how patient-reported data can feed into NLP workflows.

The \textbf{iPOF project} (investigative research into the experiences of online forum users) illustrates ethical NLP on mental health forum data (patient support groups). It developed a comprehensive ethics framework drawing on UK Government Data Ethics, BPS Internet-Mediated Research guidelines, and NHS Trust approval, published at \url{https://www.lancaster.ac.uk/health-and-medicine/research/}.

\subsubsection{Analysis of patient narratives}

\citet{han-etal-2026-dutch} extract metaphors from Dutch cancer patients' interviews and forum data using LLMs combined with human-in-the-loop validation. They connect to the broader \textbf{Metaphor Menu} project (\url{https://wp.lancs.ac.uk/melc/}) that develops multilingual conversation tools for cancer patients.

\citet{lal-etal-2024-analysing} analyse cancer narratives from Reddit to understand the emotional content around stages and treatments, providing insights into how patients articulate their health experiences in informal language.

\citet{lal-etal-2025-lens} present \textbf{LENS} (Learning Entities from Narratives of Skin Cancer), a system demonstration for extracting medical entities from cancer patient narratives in social media.

\subsubsection{Shared Decision Making (SDM)}

\citet{oueslati2026shared} study shared decision making with ethnic minorities in oncology through interviews, identifying language barriers and the need for culturally tailored information. They point to NLP tasks such as simplification, translation, and metaphor integration as important research directions.

\citet{kidanemariam2024patient} analyse clinical consultations in diabetes care using qualitative methods, showing how patients and clinicians collaborate to make care fit individual preferences and values. This forms a basis for designing conversational NLP tools that support person-centred SDM.

\bibliography{custom}


\end{document}